\pdfoutput=1

\documentclass[11pt]{article}

\usepackage[]{EMNLP2022}

\usepackage{times}
\usepackage{latexsym}
\usepackage{multicol}
\usepackage{multirow}
\usepackage{rotating}
\usepackage{booktabs}

\usepackage[T1]{fontenc}
\usepackage[utf8]{inputenc}

\usepackage{microtype}
\newcommand{\mysubsection}[1]{\vspace{0.3em}\noindent\textbf{#1}}

\usepackage{inconsolata}

\title{Understanding Interpersonal Conflict Types and their\\Impact on Perception Classification}

\author{Charles Welch $\dagger \ddagger$ \and Joan Plepi $\dagger$ \and Béla Neuendorf $\dagger$ \and Lucie Flek $\dagger \ddagger$\\
    $\dagger$ Conversational AI and Social Analytics (CAISA) Lab \\ 
    Department of Mathematics and Computer Science, University of Marburg \\ 
    $\ddagger$ The Hessian Center for Artificial Intelligence (Hessian.AI) \\
    \texttt{\{welchc,plepi,neuendob,lucie.flek\}@uni-marburg.de} 
} 

\begin{document}
\maketitle
\begin{abstract}
Studies on interpersonal conflict have a long history and contain many suggestions for conflict typology. We use this as the basis of a novel annotation scheme and release a new dataset of situations and conflict aspect annotations. We then build a classifier to predict whether someone will perceive the actions of one individual as right or wrong in a given situation. Our analyses include conflict aspects, but also generated clusters, which are human validated, and show differences in conflict content based on the relationship of participants to the author. Our findings have important implications for understanding conflict and social norms.
\end{abstract}

\section{Introduction}

Understanding social norms is critical to understanding people's actions and intents, not only for humans, but also for artificial agents. The inability for artificial agents to take these norms into account may serve as a barrier to their ability to interact with humans~\cite{pereira2016integrating}. However, perceptions of what is socially acceptable behavior vary and issues are often divisive~\cite{lourie2020scruples}. It is critical to model these differences both to build higher performing systems and better understand people~\cite{flek-2020-returning,ovesdotter-alm-2011-subjective}.

In this work we classify an individual's assessment of conflict situations using the Reddit community \texttt{r/amitheasshole} (AITA). Previous work has examined the classification of social situations involving conflict at both the individual level, and community level (for the AITA subreddit). However, it does not consider the types of conflict situations from the perspective of existing conflict-focused literature.

We explore methods of clustering descriptions of social situations involving interpersonal conflict and perform a human evaluation and analysis. After proposing a novel annotation scheme, we annotate a set of 500 conflicts with six aspects of conflict. Aspects and clusters are then used to provide an analysis of our model performance.

We address the task of predicting whether someone will perceive the actions of one individual as right or wrong in a given situation. We hypothesize that, for the prediction model, (1) higher emotional intensity will make predicting the perception of conflict more difficult, (2) when more people are involved, conflict will be harder to assess, (3) the strength of disagreement will not affect prediction difficulty, and (4) that conflict over a longer duration, involving more interference, and that are more manifest than perceived, will be easier to predict, as the additional information gives a clearer picture of the situation and points of discussion.

\section{Related Work}\label{sec:related_work}

Many classification tasks are subjective in nature. While in some cases it may help to resolve differences between annotators~\cite{hagerer-etal-2021-end}, it is often insightful to acknowledge and explore the subjectivity of labels assigned by people or groups~\cite{leonardelli-etal-2021-agreeing,sap2021annotators}. A dataset with labels from individuals, termed \textit{descriptive} annotations, will help us build models to better understand differences in people's views of socially acceptable behavior~\cite{rottger2021two}.

\citet{lourie2020scruples} first examined AITA, suggesting that the descriptive ethics contained in people's judgements could serve as a valuable resource for developing machines that can appropriately and safely interact with people. \citet{forbes-etal-2020-social} further attempted to derive rules-of-thumb from AITA to guide ethical reasoning. In contrast, our work classifies how individuals interpret these situations.

Several recent works have attempted to classify comments, or the judgement that individuals assign in their replies to posts. \citet{efstathiadis-etal-2021-explainable} examined the classification of both posts and comments on AITA, finding that posts were more difficult to classify. \citet{deCandia-2021-modeling} found that the subreddits where a user has previously posted can help predict how they will assign judgements and manually classified posts into five categories: family, friendship, work, society, and romantic relationships. More recently, \citet{botzer-etal-2022-analysis} constructed a comment classifier and used it to study the behavior of users in different subreddits.
Several of these works have examined characteristics of posts and authors and the judgements they receive, including passive voice, framing, gender, and age~\cite{zhou-etal-2021-assessing,deCandia-2021-modeling,botzer-etal-2022-analysis}.

\mysubsection{Interpersonal Conflict.} Distinctions between conflicts can be made based on who is involved. Intrapersonal occurs within oneself, while interpersonal occurs between individuals. Conflict with more people can occur within or across groups or organizations. Much research on the topic has focused on work goals and differentiates between task-related issues and those that result from differences in personality, values, or style~\cite{pinkley1990dimensions}. This work has found it useful to distinguish between conflicts concerning interpersonal incompatibilities and those that arise from the content of a task being performed~\cite{jehn1995multimethod}. Further types have been introduced, though meta-analyses have found these types to be highly correlated and thus researchers have called for improvements to how conflict is conceptualized and measured~\cite{jehn1997qualitative,korsgaard2008multilevel,bendersky2014identifying}.

\citet{barki2004conceptualizing} surveyed work on interpersonal conflict and noted that studies focused on three common attributes: disagreement, negative emotion, and interference, which correspond to cognitions, emotions, and behaviors respectively. They suggest that these aspects vary across situations and that it is important to specify the target of the conflict. They define interpersonal conflict as ``a dynamic process that occurs between interdependent parties as they experience negative emotional reactions to perceived disagreements and interference with the attainment of their goals.'' As this suggests, \textit{conflict is about perception}~\cite{hussein2019conflicts}.

\citet{korsgaard2008multilevel} referred to \citet{barki2004conceptualizing}'s three attributes as the experience of incompatibility, and suggested two additional considerations; differences in desired outcomes, behaviors, values, or beliefs, and the conflict between and among groups. \citet{bendersky2014identifying} further suggested clarifying the intensity of opposition (e.g. fight versus disagreement), specifying conflict duration, and distinguishing between perceived and manifest representations of conflict. These suggestions provided the basis of our annotation scheme described in \S\ref{sec:annotation}.

To our knowledge, no study has yet examined computational approaches to classifying the perception of social situations from the perspective of previous work on interpersonal conflict.

\section{Data}
We collected data from Reddit, an online platform with many separate, focused communities called subreddits. In particular, we use data from the AITA subreddit, where members post a description of a social situation involving an interpersonal conflict and ask other members of the subreddit if they think the author of the post is the wrongdoer in the situation or not. Others will respond saying ``you're the asshole'' (YTA), or ``not the asshole'' (NTA).
As an initial source to crawl the comments, we use the posts from \citet{forbes-etal-2020-social}. We crawl the post title together with its full text, and all the comments that contain a verdict (YTA or NTA, extracted with a list of variations). Our dataset contains 21K posts, and 364K verdicts (254K NTA, 110K YTA) written in English.
To analyze the types of conflicts, we further group posts into distinct categories as described in \S\ref{sec:clustering}.

\section{Annotation of Conflict Aspects}\label{sec:annotation}
Given the history of the typology of conflict, discussed in \S\ref{sec:related_work}, we decided to measure six aspects of conflict; (1) strength of disagreement, (2) intensity of negative emotion, (3) degree of interference, (4) duration of conflict, (5) manifestation of conflict, and (6) how many people are involved.
Aspects 1-3 correspond to the three attributes outlined by \citet{barki2004conceptualizing}, but with the view of measuring their intensity. \citet{bendersky2014identifying}'s suggestions directly inspired aspects 4 and 5 and \citet{korsgaard2008multilevel}'s suggestions about groups led to aspect 6.

The authors then annotated a sample of 25 conflicts in order to refine our task.
This process made evident how previous conflict scales were not well-equipped for our data. Our conflict situations do not always take place in work settings. The nuance of scales like \citet{jehn1995multimethod} seemed unnecessary, as conflict is assumed in our setting and as a third party, levels of intensity are less clear (e.g. how to differentiate between degrees of friction, tension, emotional conflict, and personality conflict). Longitudinal aspects also cannot often be directly determined.

With these insights we refined our annotation questions, which are provided in Appendix~\ref{sec:appendix_survey}. A subset of 500 posts corresponding to 1,653 comments from the test set were provided to annotators. Matthews correlation coefficient (MCC, \citet{matthews1975comparison}) was used to measure agreement between annotators for 100 posts and is shown in Table~\ref{tab:mcc_agreement}. We find moderate to strong agreement for most aspects with the exception of the degree of interference and whether the conflict is primarily manifest or perceived. For the non-binary aspects, we condensed labels (denoted by $\rightarrow$) and treat all labels as binary in subsequent analyses. Merged labels and label distributions are given in Appendix~\ref{sec:appendix_label_dist}.

\section{Clustering}\label{sec:clustering}

Before we acquired any annotated data, we performed an exploratory analysis to determine if there was a natural way of grouping conflicts into different types that would be useful for our analysis. We used two representations to perform clustering: situations and all text from the post (full text). Situations, as referred to in \citet{forbes-etal-2020-social}, come from the title of a Reddit post and serve as a summary of the situation described in the full post. The posts usually start with ``AITA for'', which we omit. 

We cluster posts using Louvain clustering, which maximizes the modularity of our graph~\cite{blondel2008fast}. We create a weighted graph based on each criterion, using situations or full texts as nodes. Their embeddings are obtained with Sentence-BERT (SBERT; \citet{reimers-gurevych-2019-sentence}), and use the cosine similarity, normalized to $[0,1]$ between each pair of nodes as weighted edges, resulting in two fully-connected graphs. The graphs are pruned by dropping the N\% lowest edge weights determined by the adjusted Rand index between graphs with a 10\% difference in the number of dropped edges in order to find a persistent clustering. This yields N=40\% for situations and 30\% for full texts. After clustering each had 3 clusters (see Appendix~\ref{sec:appendix_clustering}).

\begin{table}[]
    \centering
    \small
    \begin{tabular}{rc}
        \toprule
        \textbf{Conflict Aspect} & \textbf{MCC} \\
        \midrule
        Disagreement Strength & 0.39 $\rightarrow$ 0.49 \\
        Emotion Intensity & 0.33 $\rightarrow$ 0.41 \\
        Interference Degree & 0.13 $\rightarrow$ 0.20 \\
        Conflict Duration & 0.39 \\
        Manifestation or Perception & 0.10 \\
        Number of People & 0.40 \\
        \bottomrule
    \end{tabular}
    \caption{Annotator agreement using Matthews correlation coefficient for all six aspects. For non-binary aspects, the improvement after merging labels is shown to the right of the $\rightarrow$.}
    \label{tab:mcc_agreement}
\end{table}

Manually inspecting the clusters revealed that the groups differ from each other by the social relation of the author to the others in the situation, or how close the author is to others in the situation. For manual verification but also in an effort to explore possible modifications to the groupings, a subset of 100 posts were manually clustered by two of the authors, who intended to form a small number of groups based on the post title and content. While considering other possible groupings both came to the conclusion that it appears most natural to group the posts based on social relation. The events that occur in a conflict, understandably, appear strongly dependent on the relation between participants. Upon manual inspection and discussion between annotators, we find that differences arise from two sources. The first is boundaries between social relations. For instance, one annotator grouped family, romantic relationships, and best friends into one cluster, and put all other friends in a second cluster, while the other annotator put family in one cluster and all romantic relationships and friendships in a second. The second source of disagreement comes from perception of who is involved in the conflict. For instance, in one post, a person borrows an object from a family member's friend and although the family member is upset, we do not know if the friend is upset. One annotator saw this as a family conflict, while the other saw it as involving someone more distant. The ARI was 0.33 between humans, 0.38 and 0.15 between full text and humans, and 0.31 and 0.13 between humans and situations. We refer to the \textit{Family} cluster and the clusters containing \textit{Close} or more \textit{Distant} individuals in subsequent analyses. Examples from each cluster type are shown in Appendix~\ref{sec:appendix_cluster_examples}.

\begin{table*}[]
    \centering
    \small
    \begin{tabular}{rcccccccccccc}
        \toprule
         & \multicolumn{2}{c}{\textbf{Disagreement}} & \multicolumn{2}{c}{\textbf{Emotion}} & \multicolumn{2}{c}{\textbf{Interference}} & \multicolumn{2}{c}{\textbf{Duration}} & \multicolumn{2}{c}{\textbf{Manifestation}} & \multicolumn{2}{c}{\textbf{Num. People}} \\
         Diff. & \multicolumn{2}{c}{$p<0.002$} & \multicolumn{2}{c}{$p<0.02$} & \multicolumn{2}{c}{$p<0.3$} & \multicolumn{2}{c}{$p<0.04$} & \multicolumn{2}{c}{$p<0.04$} & \multicolumn{2}{c}{$p<0.007$} \\
         & Mild & Strong & Mild & Strong & Weak & Strong & Once & Longer & Perc. & Mani. & One & More \\
         \midrule
         Acc\% & 89.5 & 88.3 & 88.3 & 84.0 & 84.7 & 86.3 & 82.7 & 86.5 & 81.8 & 86.4 & 86.1 & 80.0 \\
         Micro F1\% & 70.8 & 69.5 & 70.0 & 69.6 & 56.4 & 85.5 & 68.2 & 70.7 & 51.9 & 73.7 & 73.1 & 42.5 \\
         Macro F1\% & 78.0 & 76.4 & 77.8 & 76.6 & 74.5 & 85.5 & 71.7 & 82.0 & 73.2 & 78.9 & 78.5 & 72.7 \\
         \bottomrule
    \end{tabular}
    \caption{Performance across conflict aspects (described in \S\ref{sec:annotation} and Table~\ref{tab:mcc_agreement}) for our model using the full text stratification, showing accuracy (Acc) and F1-score. Significance values for differences in model performance between each dyad are shown above, calculated with one-sided unpaired permutation tests.}
    \label{tab:conflict_aspect_results}
\end{table*}

\section{Hypotheses}

After choosing the six conflict types, we developed hypotheses about which values would be associated with conflicts whose verdicts would be most difficult for our model to predict. We hypothesized that higher emotional intensity would be more difficult, as different people may empathize differently and the classification of emotions is known to be a challenging task in itself. When more people are involved in a conflict, we hypothesized that this would be harder for our model to predict. With more involved parties, coreference resolution becomes more challenging and the interaction of more parties may make interpretation of the situational context more complex. However, we thought that the classifier would perform similarly for both mild and strong disagreements, as we did not see why this aspect by itself would make the task more or less challenging.

We predicted that it will be easier for the model to predict conflicts that occur over a longer duration, that involve more interference, and that are more manifest than perceived. First, longer duration conflicts may mean that there has been more time to accumulate information about the conflict. In our observations, it also often means that someone is repeating an action. These repeated actions and additional information may give a clearer signal of what facts will lead to a verdict. Similarly, with interference, the action is much clearer when interference is high (e.g. someone taking something away from someone, or preventing people from seeing each other). Lastly, when conflict is manifest, it means that an annotator decided the conflict was more manifest than perceived by the author. When the conflict is more perceived, the reader has to infer more from the text. For example, the author may think they did something wrong (e.g. not moving in with friend) but the author does not seem to know how the other person feels. 

\begin{table}[]
\centering
\small
\begin{tabular}{crcccc}
\toprule
& \multicolumn{1}{c}{\textbf{}} & \multicolumn{2}{c}{\textbf{Full Text}} & \multicolumn{2}{c}{\textbf{Situation}} \\
& \multicolumn{1}{c}{\textbf{}} & \textbf{F1\%}        & \textbf{Acc\%}       & \textbf{F1\%}       & \textbf{Acc\%}       \\
\midrule
\multirow{2}{*}{\rotatebox[origin=c]{90}{\parbox{3.5em}{\raggedright \citet{botzer-etal-2022-analysis}}}} & All & 72.7             & 84.9             &  70.1 & 83.2                    \\ % ft - micro 72.2, sit-micro = 68.1
& Family & 74.9 & 86.8 & 73.3 & 85.4 \\ %ft - 75.2, sit - 71.3
& Close & 72.2 & 84.4 & 67.8 & 82.2 \\ %ft - 73.7, sit - 67.5
& Distant & 71.2 & 82.2 & 68.5 & 80.8 \\ % ft - 66.3, sit - 62.9
\midrule
\multirow{2}{*}{\rotatebox[origin=c]{90}{\parbox{4.2em}{\raggedright Our Approach}}} & All & 77.2 & 87.0 & 77.4 & 87.2            \\ % ft - micro 77.3, sit - micro - 78.3
& Family & 79.0 & 88.3 & 78.7 & 88.4 \\ %ft - 78.9, sit - 79.2
& Close & 76.7 & 86.9 & 77.4 & 86.9 \\ %ft - 78.7, sit - 78.7
& Distant & 75.9 & 85.0 & 75.6 & 85.4 \\ %ft - 73.7, sit - 76.1
\bottomrule
\end{tabular}
\caption{Comparison between \citet{botzer-etal-2022-analysis} and our approach with accuracy (Acc) and macro F1-score. Results are broken down by cluster (labels from \S\ref{sec:clustering}).}
\label{tab:results}
\end{table}

\section{Perception Experiments}
We classify the perception of individuals based on their comments to posts. We concatenate the situation (post title) and comment text after filtering out any labels (e.g. YTA).
As our base model, we fine-tune SBERT on the binary task of predicting the perception of the author, given by a verdict (YTA or NTA). We also tried using this model to encode the full text to use as additional features, though we found no difference in performance over using only the comment text and situation, which often succinctly captures the event.

We compare our model to the recent work of \citet{botzer-etal-2022-analysis} JudgeBERT, which is a BERT-base~\cite{devlin-etal-2019-bert} model fine-tuned on our dataset, which is extended with a dropout layer and classification layer. JudgeBERT was evaluated in the work from \citet{botzer-etal-2022-analysis} using a dataset with collections of posts submitted between January 1, 2017, and August 31, 2019, over different subreddits. For the purpose of this work, we re-implemented JudgeBERT in order to evaluate it on our dataset. The main difference between the two models is the encoder layer, where one uses a BERT-base model, and the other one a SBERT model. We train both models for 10 epochs, using the Adam optimizer, learning rate of $1e-4$ and focal loss~\cite{8417976} to cope with class imbalance. We split our dataset into 70-20-10 for training, validation, and test, respectively. We stratify in two ways, for each clustering method.

The results are reported in Table \ref{tab:results} for both models and splits. We see that our model significantly outperforms previous work on all data,\footnote{Permutation test for full text and situations, $p<0.0001$.} with a 5 point improvement on full text F1 (macro averaged over posts, which may have multiple verdicts from different users) and 7 points on situations.

We further break down our results by conflict aspects in Table~\ref{tab:conflict_aspect_results}. We find significant differences in our model's ability to predict perception of conflicts between each aspect dyad with the exception of interference, which had a label distribution least similar to the other conditions (see Appendix~\ref{sec:appendix_judgements}). We correctly hypothesized that situations with more negative emotion would be more difficult for our classifier, though we also found this to be the case for disagreements. Further work is needed to understand the relation between disagreement strength and perception classification. We also correctly hypothesized that conflicts involving more people are more difficult for our classifier, and that stronger interference, longer duration, and primarily manifest conflicts were easier to classify, though the improvement for interference was not significant.

\section{Discussion}
Overall, our model outperforms previous work for our full data and for each cluster. As noted in \S\ref{sec:related_work}, it is important to understand the subject of the conflict, though in our work we found that this was highly coupled with the type of relation between participants. Future work may consider ways of separating these concepts.

If one considers the Family cluster as the most close social relationship, we find an indirect relationship between the closeness of participants in a conflict and the difficulty in classifying perceptions of that conflict. 

The closeness of relation to conflict participants, strength of negative emotions and opposition, duration of the conflict, manifestation, and the number of people involved all impact on our classifier's ability to classify people's perception of social norms.
These findings pertain to the understanding of conflict, behavior, and personal narratives, but may prove useful for other tasks such as argumentation, framing detection, and understanding offensive speech.

\section{Conclusions}
We developed a novel annotation scheme for aspects of conflict and built a classifier to predict individual people's perception of right and wrong. Our analysis with the aspects and generated clusters showed that the closeness in social relation between people in conflict, strength of disagreement and negative emotion, conflict duration, manifestation, and the number of people involved all impact the difficulty of predicting personal perceptions. Future work on language understanding and social norms should consider the impact of these aspects.
Our code and dataset containing 21K posts, 364K comments, two sets of cluster labels, and our 500 posts labeled with the six conflict aspects, corresponding to 1,653 verdicts is available on our GitHub.\footnote{\url{https://github.com/caisa-lab/interpersonal-conflict-types}}

\section*{Limitations}

Our experiments were performed using only English data from one subreddit discussing interpersonal conflicts. The data source conveniently provided annotated data for our application, but our findings may not fully generalize to other data sources or languages. Demographics of Reddit users are skewed toward certain populations. Similarly, we did not collect demographics of the crowd annotators, which has been shown to explain disagreements in annotation~\cite{sap-etal-2022-annotators}.

There are many modeling decisions that could lead to better performing methods. Although we explored different clustering methods and parameters in preliminary experiments, it is possible other methods and interpretation by different human annotators would lead to different cluster themes.

Our novel annotation scheme has not been thoroughly validated, and agreement for some aspects is low. The scheme and annotation instructions could be refined in future work which may lead to higher agreement, particularly for assessing interference and the manifestation of conflict.

\section*{Ethics Statement}

Better understanding social norms is important both for humans and artificial agents. Acknowledging that artificial agents could benefit from understanding that different people have different perspectives could lead to a type of author profiling task, where a model is used to predict someone's opinion of a conflict or type of conflict~\cite{rangel2013overview}. This could potentially be harmful in applications regardless of intention. We recommend against using such a model in applications where the user is unaware of data being collected about them and the purpose of collection. Even with user consent, models that misclassify user's perceptions may lead to undesired outcomes depending on the application.

\section*{Acknowledgements}
This work has been supported by the German Federal Ministry of Education and Research (BMBF) as a part of the Junior AI Scientists program under the reference 01-S20060, the Alexander von Humboldt Foundation, and by Hessian.AI. Any opinions, findings, conclusions, or recommendations in this material are those of the authors and do not necessarily reflect the views of the BMBF, Alexander von Humboldt Foundation, or Hessian.AI. 

\bibliography{rebib}
\bibliographystyle{acl_natbib}

\appendix

\section{Annotation Task}\label{sec:appendix_survey}

We recruited annotators from the crowdsourcing platform Prolific,\footnote{\url{https://www.prolific.co/}} as well as asking researchers at our university to help annotate as part of their paid working time. There were 14 annotators in total and all were required to have English fluency. All surveys included two attention check questions that provided the same options as the disagreement strength and negative emotion questions, but asked ``How should you answer this question? You should answer'', followed by one of the three options. All annotators passed all attention checks.
Annotators were asked the following six questions for each Reddit post and additional details on how the labels should be used:

\begin{table*}[]
    \centering
    \small
    \begin{tabular}{lcccccccccc}
        \toprule
        Cutoff \% & 0 & 10 & 20 & 30 & 40 & 50 & 60 & 70 & 80 & 90 \\
        \midrule
        Number of Situation Clusters & 4 & 3 & 4 & 3 & 3 & 4 & 4 & 4 & 4 & 4 \\
        Situation ARI & - & 0.44 & 0.47 & 0.46 & 0.93 & 0.57 & 0.45 & 0.49 & 0.91 & 0.85 \\
        Number of Fulltext Clusters & 3 & 3 & 3 & 4 & 3 & 5 & 8 & 18 & 49 & 165 \\
        Full Text ARI & - & 0.60 & 0.92 & 0.91 & 0.75 & 0.72 & 0.74 & 0.89 & 0.81 & 0.65 \\
        \bottomrule
    \end{tabular}
    \caption{The resulting number of clusters using Louvain for different graph representations, and cutoff percentages. ARI denotes the adjusted rand index between the listed cutoff percentage and 10\% less.}
    \label{tab:louvain_results}
\end{table*}

\begin{table*}[]
    \centering
    \small
    \begin{tabular}{cccccccccccc}
        \toprule
         \multicolumn{2}{c}{\textbf{Disagreement}} & \multicolumn{2}{c}{\textbf{Emotion}} & \multicolumn{2}{c}{\textbf{Interference}} & \multicolumn{2}{c}{\textbf{Duration}} & \multicolumn{2}{c}{\textbf{Manifestation}} & \multicolumn{2}{c}{\textbf{Num. People}} \\
         Mild & Strong & Mild & Strong & Weak & Strong & Once & Longer & Perc. & Mani. & One & More \\
         \midrule
         33.0 & 67.0 & 35.7 & 64.3 & 35.3 & 64.7 & 48.3 & 51.7 & 33.7 & 66.3 & 72.0 & 28.0 \\
         \bottomrule
    \end{tabular}
    \caption{Label distribution for merged label values resulting from human annotation of 500 posts.}
    \label{tab:label_distribution}
\end{table*}

\begin{enumerate}
    \item How strong is the disagreement or opposition? \textbf{Labels:} (Mild, Strong, Intense) with Strong and Intense merged. \textbf{Additional details:} You should consider how significant the event seems to the author. For example, a conflict over who should clean the dishes may seem mild, whereas a conflict over divorce may seem intense. However, if the author describes the conflict over dishes as a fight that is causing irreparable damage to the relationship, it may be strong or intense.
    \item How intense are the negative emotions? \textbf{Labels:} (Mild, Strong, Intense) with Strong and Intense merged. \textbf{Additional details:} Use the mild label when emotions are weaker, or it is not clear if they are there at all. Use the strong and intense labels to differentiate between situations where you perceive stronger emotions from the participants.
    \item How much is one person interfering with what another wants to or can do? \textbf{Labels:} (Not at all, Somewhat, Strongly) with Not at all and Somewhat merged. \textbf{Additional Details:} If someone clearly cannot do what they would like and that is the subject of the conflict, then the interference is strong. If there is a disagreement, but parties can still take whichever action they desire, then there is no interference (e.g. telling someone not to do their homework but not stopping them from doing it). If there are alternatives or possibility for some degree of compromise then there is some interference (e.g. a tenant is upset that they cannot pay rent in two parts, landlord gives several alternatives), but if the restricted party is clearly opposed to all options then the interference is still strong (e.g. daughter is not allowed to go to boyfriends house).
    \item What is the duration of the conflict? \textbf{Labels:} (One-time incident, Longer) \textbf{Additional Details:} Additional Details: If someone describes a specific incident that occurred at one point in time then it is a one-time incident (e.g. posting something rude one time on Facebook, not wanting sibling to take over a family vacation with her plans). If the author explicitly states that something is an ongoing conflict over multiple days (or longer), or if it can be reasonably inferred that a conflict spans multiple days (e.g. “every time I talk to my parents we have this problem”), then the conflict is longer term.
    \item Has the conflict primarily manifested in what someone has said or done, or is the conflict primarily perceived by the author? \textbf{Labels:} (Manifest, Perceived) \textbf{Additional Details:} Additional Details: A conflict can become manifest, for example, in the form of fights, arguments, telling someone something, or taking something, whereas the perception of conflict happens inside someone's head (e.g. someone thinks of themselves as rude/mean/unfair, but we do not know if another party has this same perception because we do not know what they have said or done or if they are aware of or have engaged in the same events as the author). For example, the author feels bad for not texting his parents back quickly. If we have no evidence that this is causing problems between them or that the parents have a problem with this then it is perceived. Sometimes there are small manifestations, but the conflict is still mostly perceived. For instance, the author is blocked on Facebook for not inviting a friend to a party, but the author does not seem to engage with the other person or understand why this is a conflict. In this case it is primarily perceived by the other person.
    \item Who else is directly in conflict with the author? \textbf{Labels:} (One person, Multiple people) \textbf{Additional Details:} Additional Details: A conflict with multiple people should only count people engaging with or contributing to the conflict. For example, if A tells B to shave their beard and C gets mad at B for doing so, B and C are in conflict but as long as A does not engage, they should not be considered to be part of the conflict and so this would be a one person conflict.
\end{enumerate}

\section{Clustering} \label{sec:appendix_clustering}

When clustering, we first determined how many edges from our fully-connected graphs to drop. This was determined using the adjusted rand index between 10\% differences. Further threshold values, ARI, and resulting cluster numbers are provided in~\ref{tab:louvain_results}. Although we do use a cutoff of 30\% for full texts, which has 4 clusters, one of these clusters contained only 25 posts, so we removed it. We experimented with K-means in preliminary experiments but found that it had lower agreement with human clusters and clusters seemed less clear.

\section{Judgements Across Aspects}\label{sec:appendix_judgements}
We also find that the types of judgements in our sample vary significantly across each aspect of conflict. The difference in the distribution of NTA and YTA labels between each dyad shown in Table~\ref{tab:conflict_aspect_results} is statistically significant using Fisher's exact test. In the difference for disagreement ($p<0.004$), \textit{Strong} contains an 11\% higher ratio of YTA/NTA judgements. For emotion ($p<0.02$), this difference was 9\%. Interference ($p<0.001$) had the highest difference of 78\%, with more YTA judgements when the degree was \textit{Strong}. For duration ($p<0.001$), \textit{One-time incidents} had a 13\% higher ratio. Manifestation of conflict ($p<0.0003$) showed a 13\% higher ratio when conflict was more manifest than perceived. Lastly, when only one person was involved ($p<0.03$), the ratio of YTA/NTA was 11\% higher. All ratios skew toward more NTA, as this is the overall bias of the dataset, and all differences in ratio are calculated as absolute differences of YTA/NTA between values of an aspect.

\section{Label Merging and Distribution}\label{sec:appendix_label_dist}

As discussed in \S\ref{sec:annotation}, we merged labels for aspects that had more than two labels. The \textit{strong} and \textit{intense} labels for the negative emotion and disagreement aspects were merged into one \textit{strong} category. The \textit{lesser} and \textit{none} labels for the degree of interference were merged into \textit{mild}. Other labels were already binary and were unchanged. The resulting distribution is shown in Table~\ref{tab:label_distribution}.

\section{Cluster Examples}\label{sec:appendix_cluster_examples}

Two examples of posts belonging to each of the clusters are shown in Table~\ref{tab:cluster_examples}. Clusters were obtained using the full text.

\begin{table*}[]
    \centering
    \small
    \begin{tabular}{p{.95\linewidth}}
        \toprule
        \multicolumn{1}{c}{\textbf{Family}} \\
        \midrule
        \textbf{Situation:} Helping my sister take my parents cat \\
        \textbf{Full Text:} Some context: my sister raised a litter of kittens from 4 days old, and my parents decided to keep one of them. We'll call him F.  F wasn't learning to stay off counters, so my mother put a shock collar on him. My sister learned of on her birthday, and is vehemently against it, saying that it's cruel, and that cats don't learn like dogs do.  Last night, I helped my sister sneak him out of the house, to her college dorm. AITA? \\
        \\
        
        \textbf{Situation:} Not watching horror films with my husband \\
        \textbf{Full Text:} I really don't like horror movies. I dislike gore and loud noise out of nowhere shock tactics especially, but I also have a tendency to get nightmares from movies that don't have those issues. I don't enjoy being scared. Plot holes also stick out like a sure thumb in horror to me.  I will try movies on occasion if he really wants me to see them and he says it isn't a gore/shock tactic movie, but it takes a lot of pleading on his part. I almost never enjoy them and generally my reaction is that it was okay/fine, wouldn't watch it again.  I watch things I want to see but he wouldn't enjoy separately. I ask him to watch things that I think he will actually like sometimes and he always does. He often watches horror after I go to bed. The things we watch together are things we are both agreeable to. We watch at home.   I only wonder if I'm the asshole because it seems common for couples to trade off who picks movies. \\
        
        \midrule
        \multicolumn{1}{c}{\textbf{Close Relationships}} \\
        \midrule
        \textbf{Situation:} Feeling abandoned by all my friends after a break up \\
        \textbf{Full Text:} Well, long story short , i had no friends until I met this girl which I dates for about a year, she included me in his close circle of friends, and I thought they like me for who I was, not only because we were dating. Oh, well, I was wrong. we break up, and now none of my supposed friends talk to me, no one wants to hang out, and when I pointed that out to the one of them that I feel the more trust via text message, he just call me an asshole, but whatever, that doesn't change the facts that I'm now as alone as I started. Roast me reddit \\
        \\
        
        \textbf{Situation:} Not attending my friend's debut \\
        \textbf{Full Text:} She already placed me on a list where they call people up to give gifts ad stuff without eve asking beforehand if I'll be able to attend. I feel like a real asshole right now because 18th birthdays only happen once in a lifetime ad I wasn't there to celebrate with her when she was expecting me because I needed to attend a birthday for my uncle who was released out of prison. On the other hand, I do feel a bit angry that she listed me before asking. Now everyone has cards with my name on them, ad whoever is attending will expect me to join as well. I feel some conflict. She didn't even tell me the address, she just told me that I'm invited and my name is on the card and I need to give her a gift. She seemed really disappointed days ago when I told her that i could't attend. Stopped talking to me. Didn't even look at me. Tried texting my other friends who were invited but didn't respond. Too busy partying. I have a feeling that people will think of me as a shitty friend and that I'm no good. So, AITA? \\
        
        \midrule
        \multicolumn{1}{c}{\textbf{Distant Relationships}} \\
        \midrule
        \textbf{Situation:} Leaving low tips \\
        \textbf{Full Text:} So there was an event at a bar/club I bought a ticket for online, *pre-paid* - but when I got there, even though I had a ticket, they were unable to let me in due to "max capacity".  I mean, normally I don't take it to heart and either wait or find somewhere else, but this was something that I paid for, so I figured it's not fair since I technically paid to be part of that `capacity'. There were a few others in the same boat as me who they had to do that to who were also frustrated.  Eventually I got in, but I was super aggravated because I ended up missing over an hour of the event because of this, and while I was able to eventually enjoy my night I found myself leaving low tips, since I was quite livid (and felt I lost some of my money's worth).  Later on I felt kind of bad because I realized it's probably not the bartenders' faults. AITA though? \\
        \\
        
        \textbf{Situation:} Getting mad at an elderly co-worker for always getting my name wrong \\
        \textbf{Full Text:} Ok, so i work for a store and one of the employees is this elderly man, about 71 or so. Now, he always gets my name wrong.  He always greets me as "Eddie". My name is nowhere close to Eddie. There is no Eddie anywhere in the store. I'm the only one he calls by the wrong name.   "How goes it, Eddie?"  "Eddie, why are you stacking those like that?"  "Eddie, that's not how you use the coffee machine!"  At first, i let it slide because i just figured he was senile and didn't know who i was. I corrected him, he called me by my name for about a day. The next day, he kept calling me Eddie.  TBH i wouldn't mind it, this guy is kind of a prick. He isn't above me in terms of position, we hold the same position. He's not a manager or anything. But he corrects me on every little thing. Even though i'm doing it the way the boss told me.  My first day stocking shelves, i was apparently putting the stuff up wrong. I was putting top shelf items on the bottom shelf. The manager corrected me. While the manager is trying to show me the right way, he shouts across the room. "Now Eddie! I know you got more sense then that! Put that stuff on the bottom shelf where it belongs!"  The manager was already telling me how, but he chose to embarrass me in front of the entire store.  He makes fun of me for being on a diet. I got a salad for lunch and  he started mocking me "Hell, Eddie, that's not enough to even keep a damn bird alive!"  So, i finally snapped. I shouted at him that my name wasn't Eddie.   "My God, My name's not Eddie! Jesus, if you're gonna act like you run this place, at least get my fucking name right!"  Everyone in the store was staring at me and i feel kind of guilty. But was i truly the a-hole in this situation? \\

        \bottomrule
    \end{tabular}
    \caption{Two examples of post situations and full text for each of the three clusters (manually labeled, but automatically clustered using the full text).}
    \label{tab:cluster_examples}
\end{table*}

\end{document}